\documentclass[conference]{IEEEtran}
\IEEEoverridecommandlockouts
\usepackage{times}

\usepackage[numbers]{natbib}
\usepackage{multicol}
\usepackage[bookmarks=true]{hyperref}
\usepackage{xcolor}
\usepackage{graphicx}
\usepackage{booktabs}
\usepackage{subcaption}
\usepackage{amsfonts}

\usepackage{fancyhdr} 

\begin{document}
\title{Anticipation through Head Pose Estimation:\\ a preliminary study}


\author{
\authorblockN{Federico Figari Tomenotti}
\authorblockA{MaLGa-DIBRIS, Universit\`a degli Studi di Genova\\
federico.figaritomenotti@edu.unige.it}

\and
\authorblockN{Nicoletta Noceti}
\authorblockA{MaLGa-DIBRIS, Universit\`a degli Studi di Genova\\
nicoletta.noceti@unige.it}
}

\pagestyle{fancy} 
\fancyhead{} 
\renewcommand{\headrulewidth}{0pt}  
\fancyhead[L]{\fontsize{9}{11}\selectfont \textit{Accepted at the workshop on advancing Group Understanding and robots' adaptive behavior (GROUND)\\
Robotics Science and Systems (RSS) Conference, 2024.}} 


%

\maketitle

\thispagestyle{fancy} 

\begin{abstract}
The ability to anticipate others' goals and intentions is at the basis of human-human social interaction. Such ability, largely based on non-verbal communication, is also a key to having natural and pleasant interactions with artificial agents, like robots. 
In this work, we discuss a preliminary experiment on the use of head pose as a visual cue to understand and anticipate action goals, particularly reaching and transporting movements. By reasoning on the spatio-temporal connections between the head, hands and objects in the scene, we will show that short-range anticipation is possible, laying the foundations for future applications to human-robot interaction.\\

\end{abstract}

\IEEEpeerreviewmaketitle

\section{Introduction}\label{sec_introduction}
A key element of natural human-human interaction is the ability to anticipate humans' goals and intentions \cite{pezzulo2012proactive}. The same ability is paramount in different application domains -- ranging from gaming to domotics and home assistance, to robotics. In the latter, in particular, anticipation abilities may enable robots to seamlessly interact with humans in shared environments, enhancing safety, efficiency and fluidity in Human-Robot Interaction scenarios \cite{huang2016anticipatory}. 

Over the last years, the importance of leveraging non-verbal cues for understanding humans' intentions has been well assessed \cite{7151799,duarte2018action}. The interesting work in \cite{moller2015effects} for instance leverages the fact that anticipatory eye movements occur during both action observation and action execution. More in general, the focus of attention of a subject involved in an interaction can provide valuable insights about their intentions, enabling the identification of the next object they will interact with
-- when interacting with the environment -- 
or the person they will engage with -- when involved in group interaction. 
In this study, we explore the feasibility of using video-based head orientation -- a proxy of human gaze -- for anticipation, with particular attention to reaching and manipulation actions.\\

In the Computer Vision literature, anticipation often refers to the task of predicting the occurrence of future actions in long-term activity recognition \cite{gong2022future,zhang2024object,Roy2021ActionAU}. In the last years, a growing interest has been devoted to the specific domain of egocentric vision (see for instance \cite{girdhar2021anticipative,furnari2023streaming}). In our work, we are focusing on short-term anticipation, more relevant for seamless human-human interaction.  For this reason, approaches based on third-person viewpoints are of interest. For instance \cite{zatsarynna2023action} aims at the anticipation of next action of 1 sec, by exploiting the high-level goal pursued in the video. In \cite{wang2023memory} it is presented a memory-based approach using Transformer architectures to address the online action detection and anticipation tasks. The methods in \cite{vondrick2016anticipating,DBLP:conf/bmvc/GaoYN17a} learn how to anticipate by predicting the visual representation of images in the future.\\
Differently from these approaches, in our work, we are interested in understanding and quantifying low-level features for determining the ability to predict in advance some cue of the current action goal. In this sense, our work is inspired by the concept of \emph{effect anticipation} \cite{kunde2007no}. \\
More in detail, we hypothesize we can use the 3D Head Direction as a proxy of the gaze, and that by deriving simple visual geometrical cues in an unsupervised way -- connecting the head and hands of a subject with the elements in the environment -- we can anticipate the goal of an action in terms of next active object or target position (when the movement involves a change in location of objects). The goal is achieved using object and human pose detectors, deriving the 3D head pose and reasoning on the interaction between the human and the scene and how it evolves over time.
To test this hypothesis, we conducted preliminary experiments using a private dataset including videos of different subjects sitting in front of a table and instructed to perform a series of reaching and manipulation actions. We show that the Head Direction can effectively and efficiently anticipate the actions' goal in such a scenario. \\
We believe the methodology discussed in this contribution may be profitably adopted as a building block for supporting HRI applications. Indeed, a robot might be engaged in smooth interactions with one or more people, favoured by the ability to anticipate the intentions of others. This opportunity will be briefly outlined in the conclusions.

\section{Method}\label{sec_method}
\begin{figure*}
\begin{subfigure}{.33\textwidth}
  \centering
  \includegraphics[width=.95\linewidth]{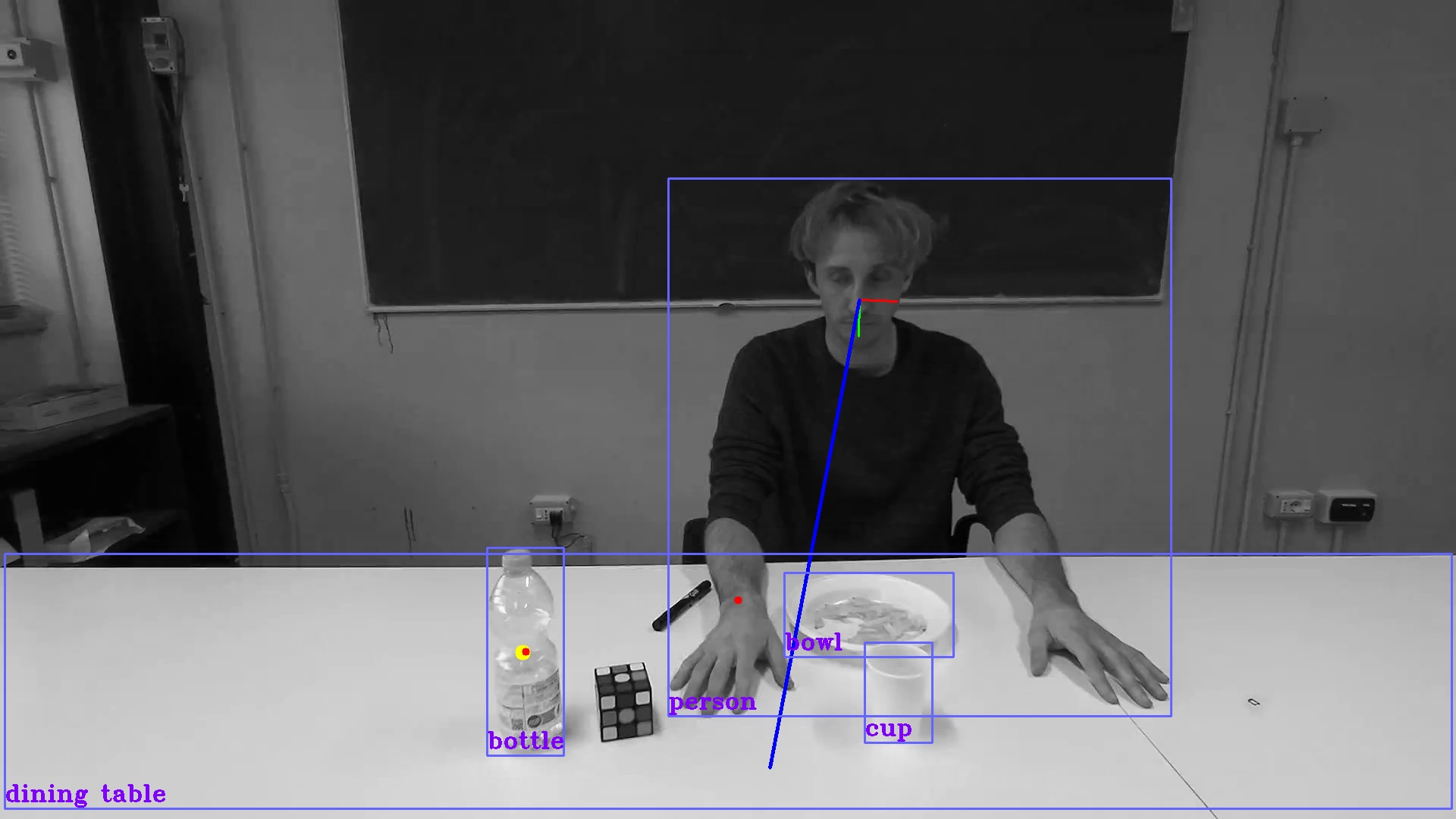}
  \caption{}
  \label{figSub_sfig1}
\end{subfigure}%
\begin{subfigure}{.33\textwidth}
  \centering
  \includegraphics[width=.95\linewidth]{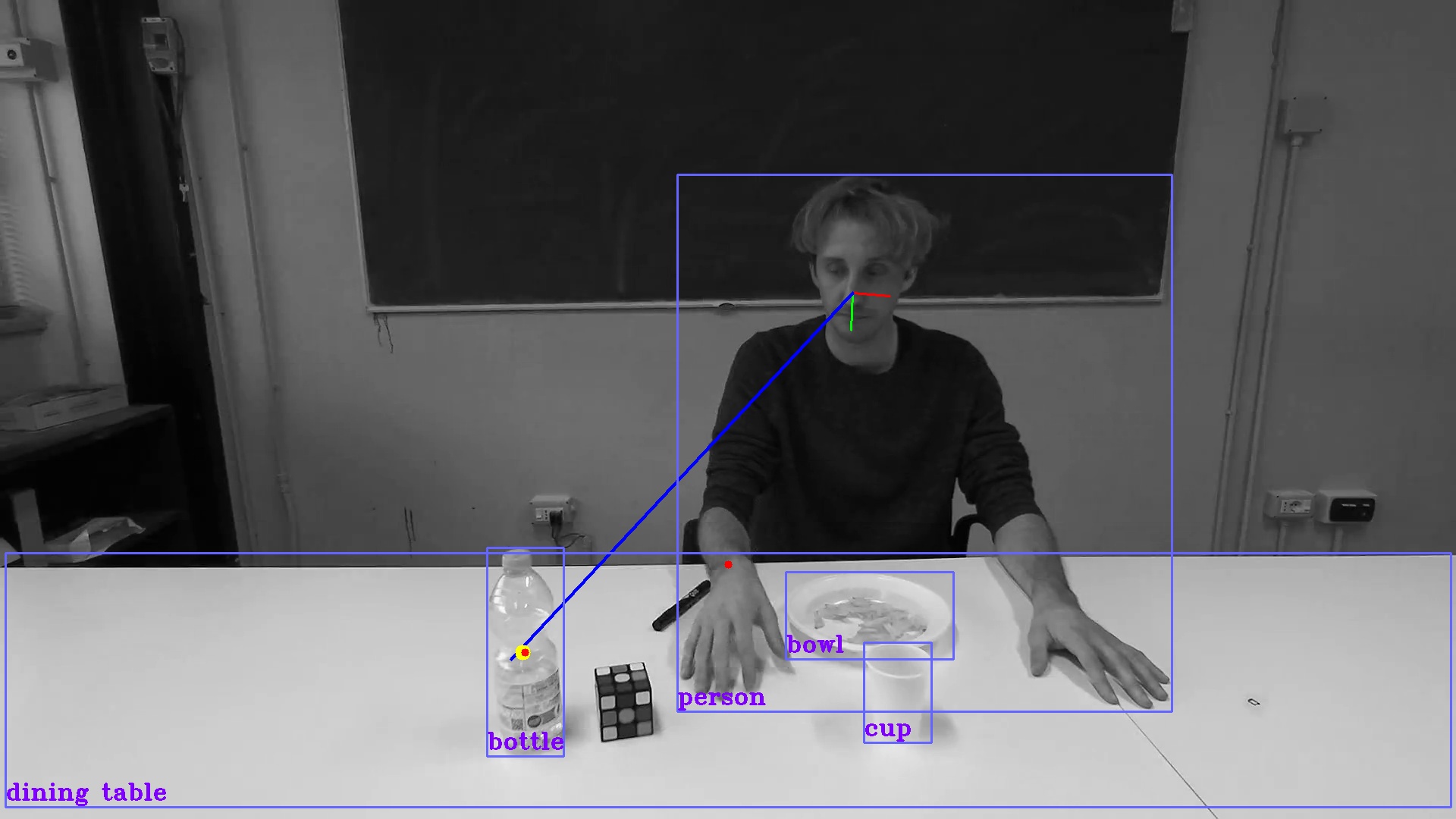}
  \caption{}
  \label{figSub_sfig2}
\end{subfigure}
\begin{subfigure}{.33\textwidth}
  \centering
  \includegraphics[width=.95\linewidth]{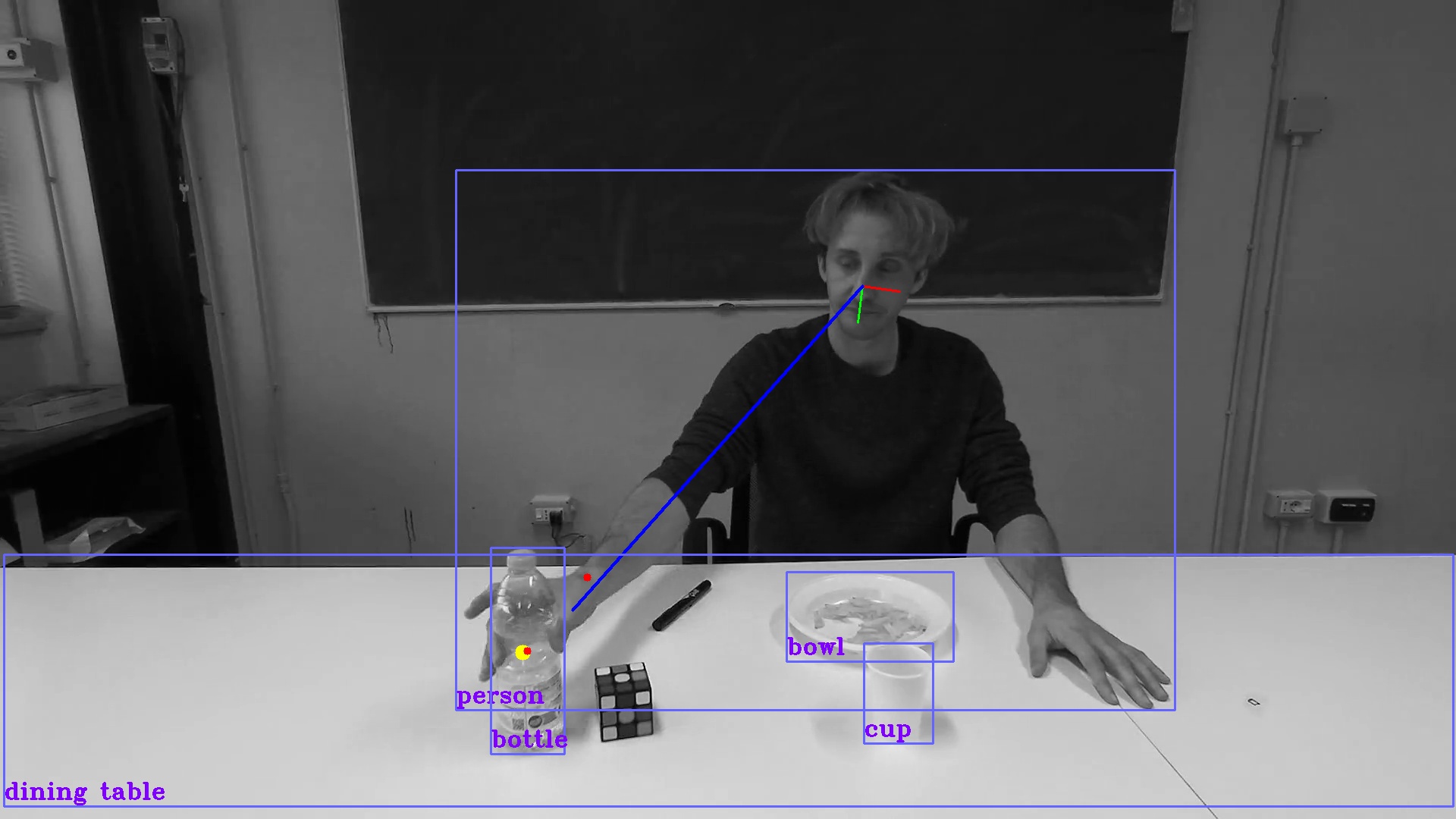}
  \caption{}
  \label{figSub_sfig3}
\end{subfigure}

\begin{subfigure}{.33\textwidth}
  \centering
  \includegraphics[width=.95\linewidth]{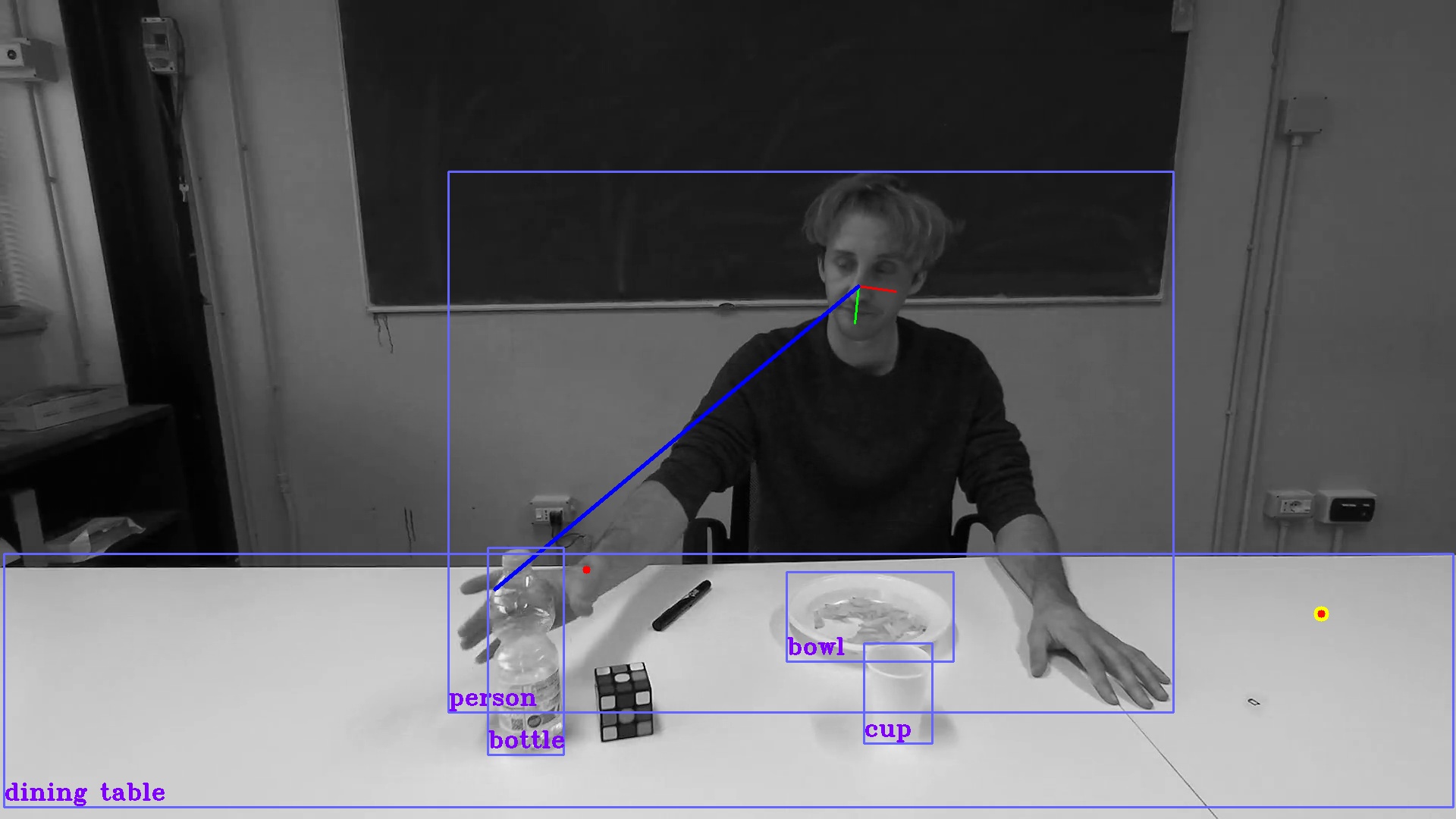}
  \caption{}
  \label{figSub_transport_frames1}
\end{subfigure}%
\begin{subfigure}{.33\textwidth}
  \centering
  \includegraphics[width=.95\linewidth]{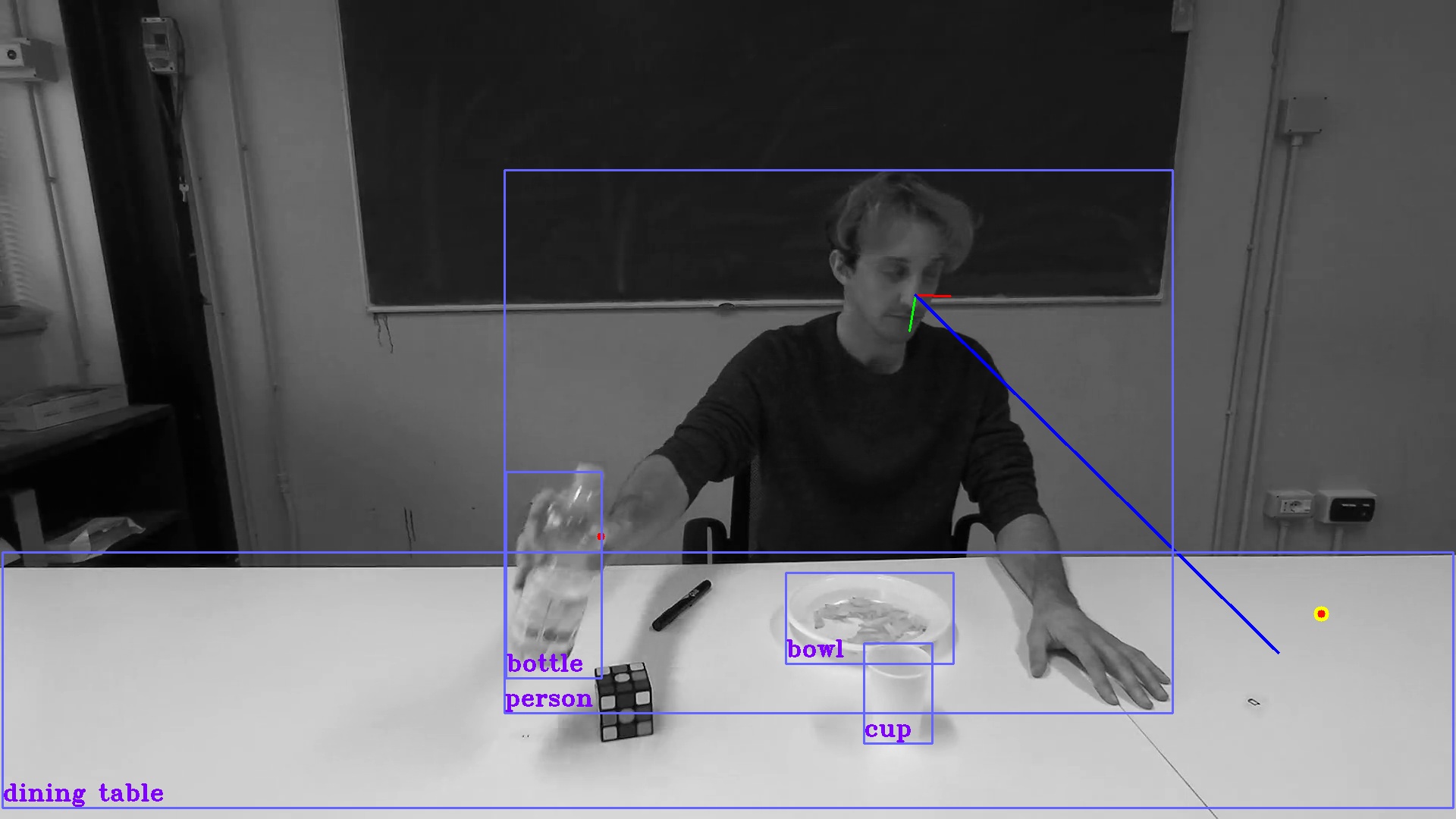}
  \caption{}
  \label{figSub_transport_frames2}
\end{subfigure}
\begin{subfigure}{.33\textwidth}
  \centering
  \includegraphics[width=.95\linewidth]{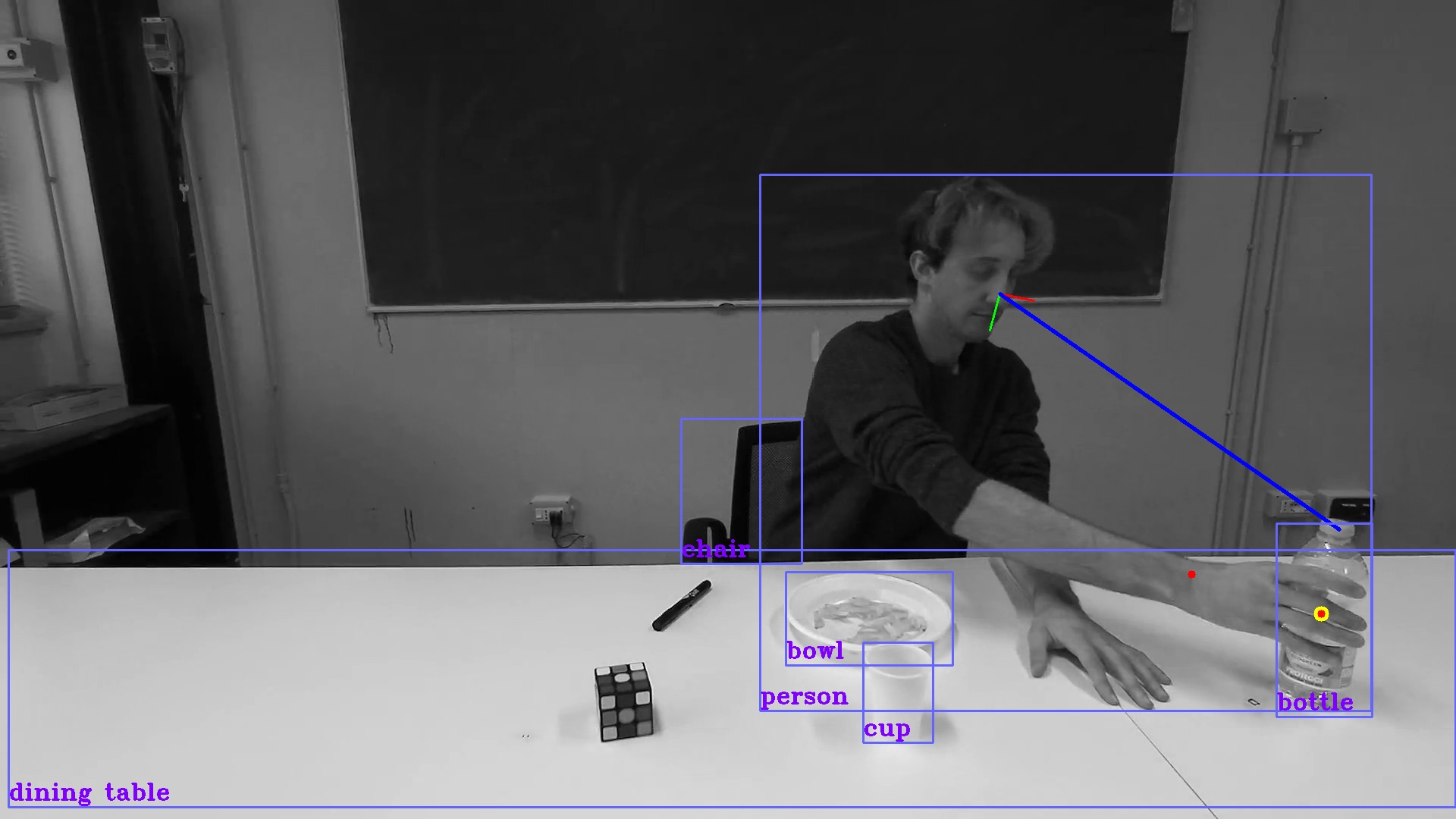}
  \caption{}
  \label{figSub_transport_frames3}
\end{subfigure}
\caption{(a,b,c): reaching from \emph{transporting action}.\ref{figSub_sfig1} The head starts to move. \ref{figSub_sfig2} The head projection onto the table reached the target. \ref{figSub_sfig3} The hand reached the target after more than 10 frames (1/3 of a second).\\\hspace{\textwidth}
(d,e,f): transporting from \emph{transporting action}. \ref{figSub_transport_frames1} The head and hand are aligned. \ref{figSub_transport_frames2} The head projection onto the table reached the target at the very beginning of the hand movements. \ref{figSub_transport_frames3} The hand reached the target. \\\hspace{\textwidth}
In light blue/violet bounding boxes and annotations are shown. The image is in black-white only for clarity.}
\label{fig_frames_example}
\end{figure*}

The methodology takes as input RGB image sequences and through a multi-stage pipeline outputs different types of information: human pose, head pose, and objects in the scene. 
Thanks to these cues, we can retrieve the time at which the hand of a person approaches the target position and the time in which the human head points towards the same target/object. In this context we use the head pose as a proxy for a proper gaze measure. We do not think this can be a weakness of the method but rather an advantage, because we can estimate head position in a broader range of situations: far from the camera, with small occlusions on the eye region and using a camera instead of glasses.

More specifically, the pipeline starts with a pose detector (we adopt Centernet \cite{zhou2019objects}, but other solutions can be used) and an object detector, for instance, YOLO (and more specifically YOLOv8 \cite{YOLO} in our case). While the first provides a list of 25 3-dimensional key points on the body, the latter detects the bounding boxes of objects in the scene and determines their classes.

Then we exploit the 5 facial key points (on eyes, nose and ears, with their detection confidence) to estimate the 3D head pose with HHP-Net \cite{Cantarini_2022_WACV, Tomenotti2024HeadPE}.
The head pose is represented with 3 angles which define the head orientation in the 3D space: yaw, pitch and roll. On the image plane, we projected them following the Tait-Bryan angles convention as explained in \cite{Tomenotti2024HeadPE} and representing them according to 3 versors parallel to the head axis; in Fig. \ref{fig_frames_example} they are represented in red, green and blue.
We hypothesise the consistency between gaze and head position and we project the field of view of the detected person on the working table (see Fig. \ref{fig_frames_example}, where we visualize this operation by extending the blue component to reach the table).

Our analysis is based on three relevant moments in simple human-object interactions involving reaching and transporting actions, i.e. (I) the moment when a person looks at the target, it could be an object for a reaching or a final position for the transport (we will refer to as \emph{gazing\_target\_time}), (II) the one where the person touches an object with the hand at the end of a reaching (we will call \emph{touching\_object\_time}), and (III) the one where an object is placed in the target position for transports (called \emph{target\_object\_time}). The hypothesis is that \emph{touching\_object\_time} and \emph{target\_object\_time} always follow \emph{gazing\_object\_time}. 

To detect these moments we decided to use, as a reference measure, the distances between the hand of the person and the target object/position. As shown in Fig. \ref{fig_frames_example}, the position of the object at time t $\mathbf{P}_O^t$ is represented by the centroid of its bounding box resulting from the object detector. We keep track of the current position of the object (in red) and the initial one (in yellow). The current position of the hand $\mathbf{P}_H^t$ is approximated by the wrist key point (also in red). To approximate what we call "the position" of the gaze $\mathbf{P}_G^t$, we consider the distance between the final point of the blue segment (with the origin on the head) and the target object. Finally, the target position for transporting movements $\mathbf{P}_T^t$ (again in yellow, on the table) is fixed and available as prior knowledge.

To summarize, we detect the three relevant moments considered in our analysis as follows:
\begin{itemize}
\item \emph{gazing\_target\_time} corresponds to the instant t where the distance between $\mathbf{P}_G^t$ and $\mathbf{P}_O^t$ or $\mathbf{P}_T^t$ is minimized
\item \emph{touching\_object\_time} is the instant with minimum distance between $\mathbf{P}_H^t$ and $\mathbf{P}_O^t$
\item \emph{target\_object\_time} is detected when the distance between $\mathbf{P}_O^t$ and $\mathbf{P}_T^t$ is minimized
\end{itemize}

The \emph{anticipation} is then estimated as \emph{touching\_object\_time}-\emph{gazing\_target\_time} (where target is the object) for reaching movements, and \emph{target\_object\_time}-\emph{gazing\_target\_time} (where target is the final position).

\begin{figure*}[p]
\begin{subfigure}{.33\textwidth}
  \centering
  \includegraphics[width=.95\linewidth]{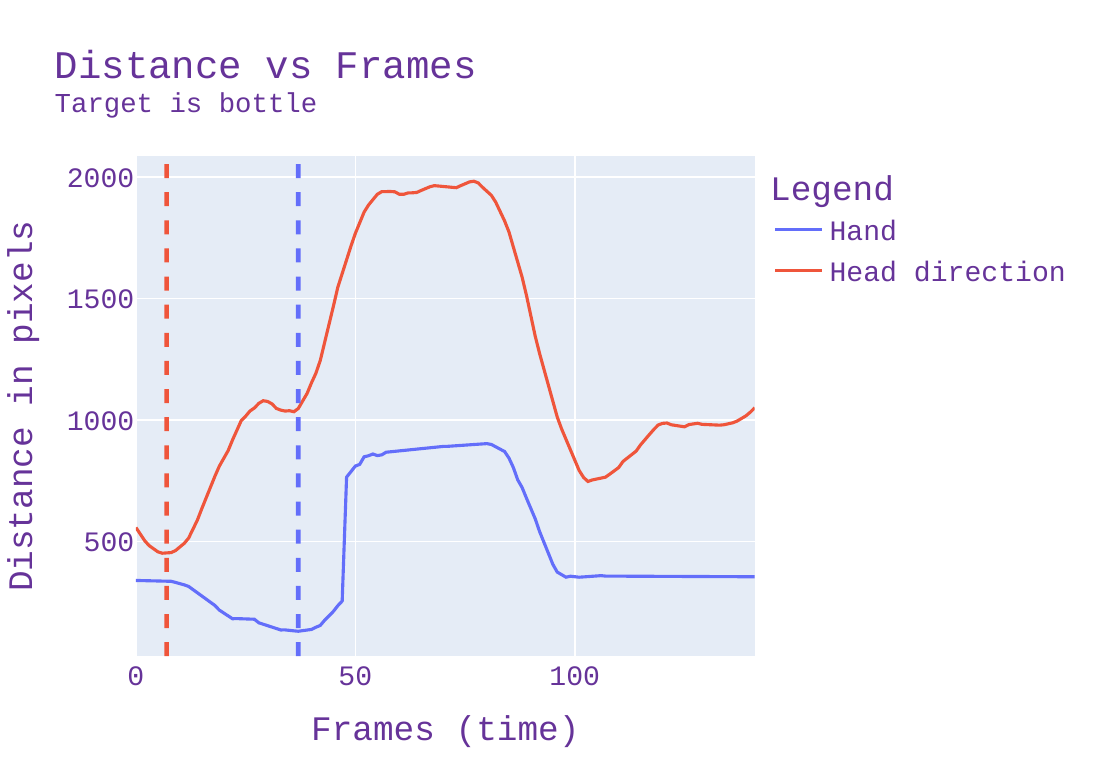}
  \caption{}
  \label{figSub_transport_bottle_1}
\end{subfigure}%
\begin{subfigure}{.33\textwidth}
  \centering
  \includegraphics[width=.95\linewidth]{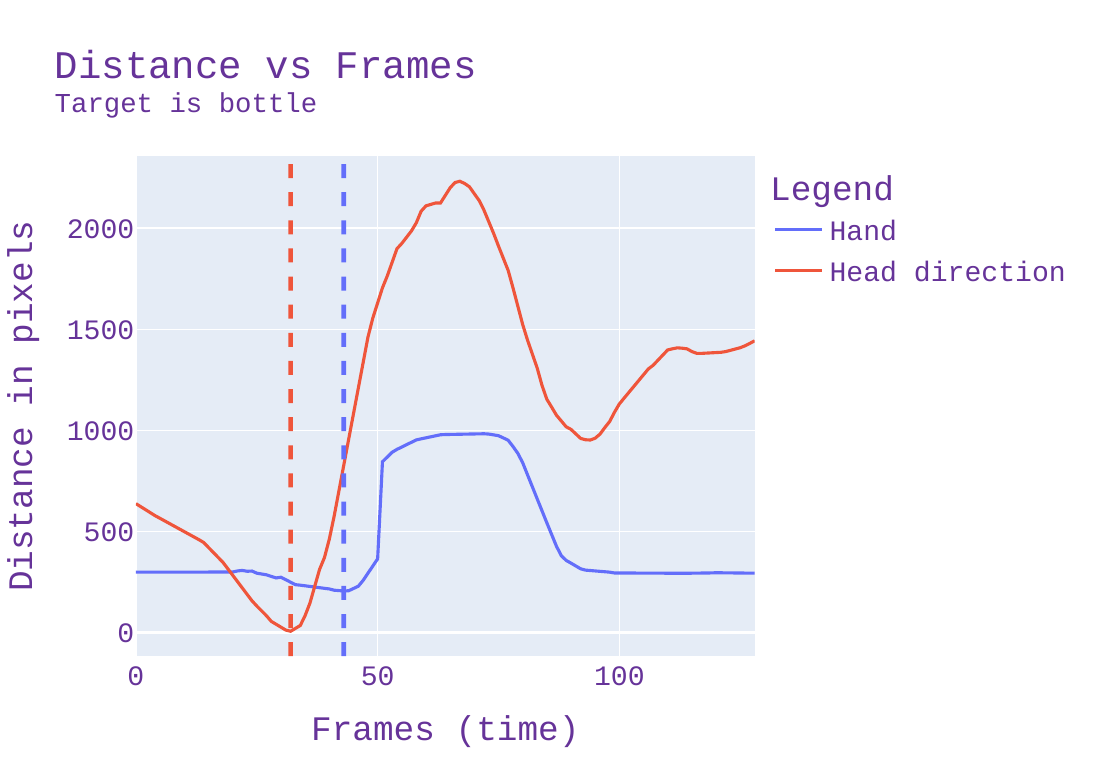}
  \caption{}
  \label{figSub_transport_bottle_2}
\end{subfigure}
\begin{subfigure}{.33\textwidth}
  \centering
  \includegraphics[width=.95\linewidth]{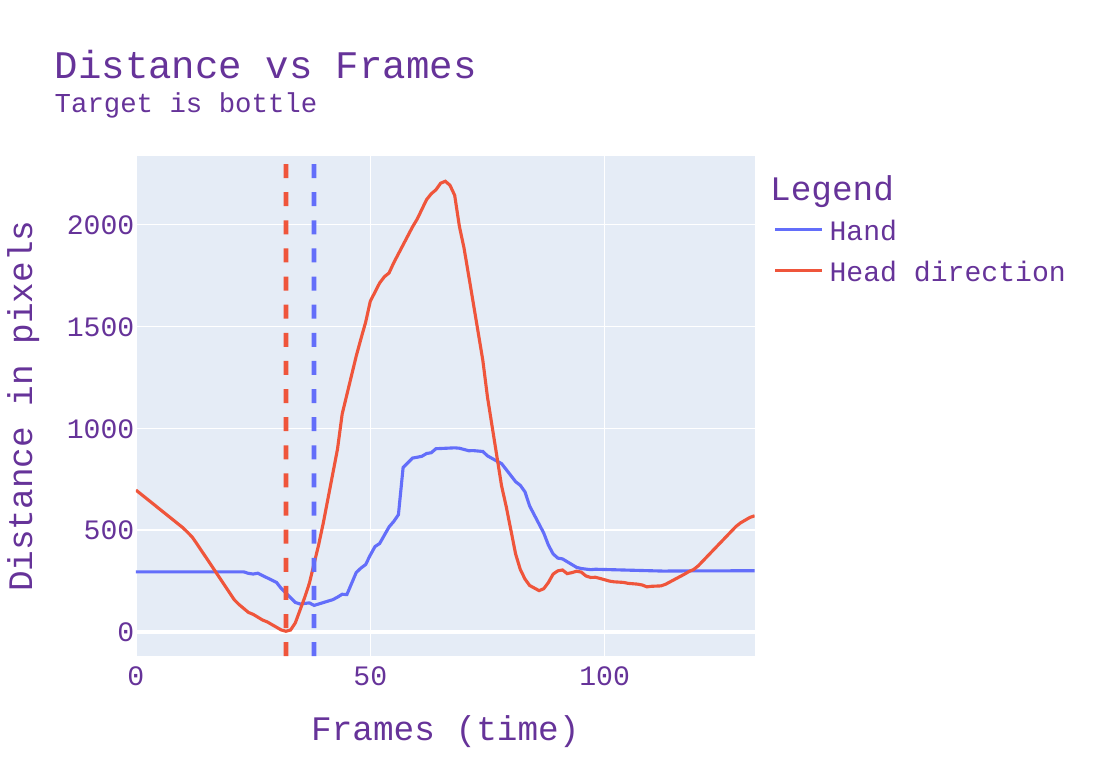}
  \caption{}
  \label{figSub_transport_bottle_3}
\end{subfigure}
\caption{Reaching the bottle in a \emph{transporting action}. The vertical lines indicate the \emph{gazing\_target\_time} (red) and the \emph{touching\_object\_time} (blue). The colour code is the same for the time-lines and the vertical lines.}
\label{fig_transport_bottle}
\end{figure*}

\begin{figure*}[p]
\begin{subfigure}{.33\textwidth}
  \centering
  \includegraphics[width=.95\linewidth]{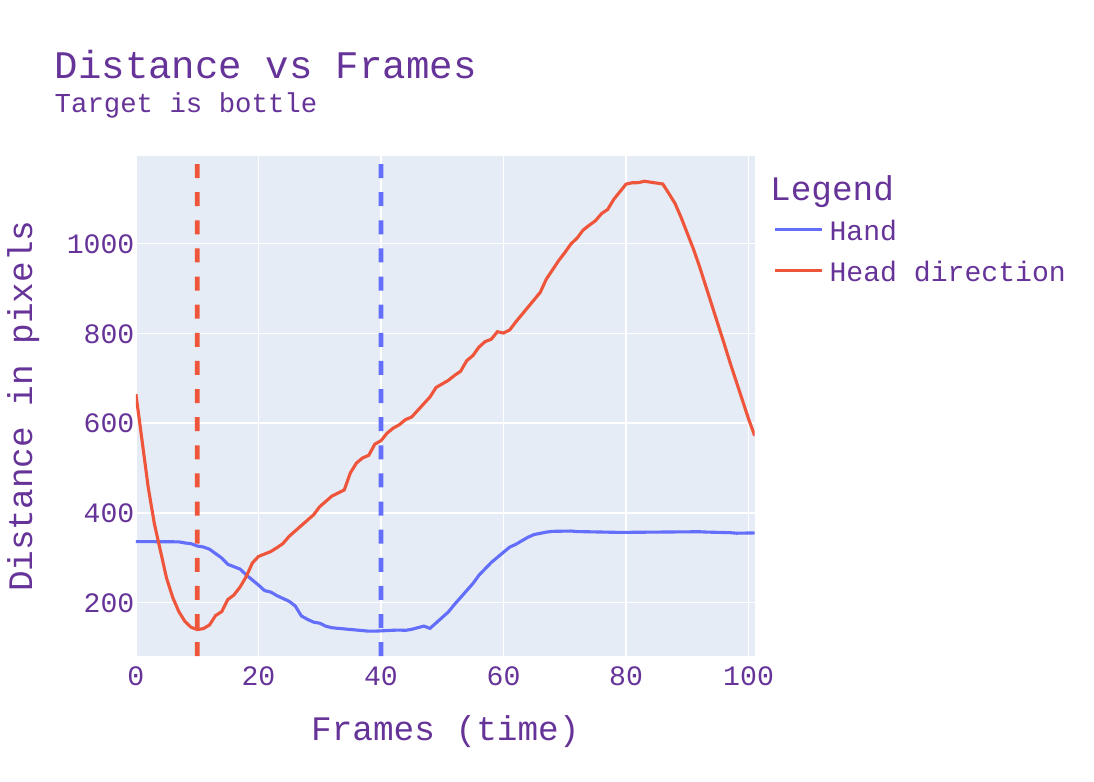}
  \caption{}
  \label{figSub_touch_bottle_1}
\end{subfigure}%
\begin{subfigure}{.33\textwidth}
  \centering
  \includegraphics[width=.95\linewidth]{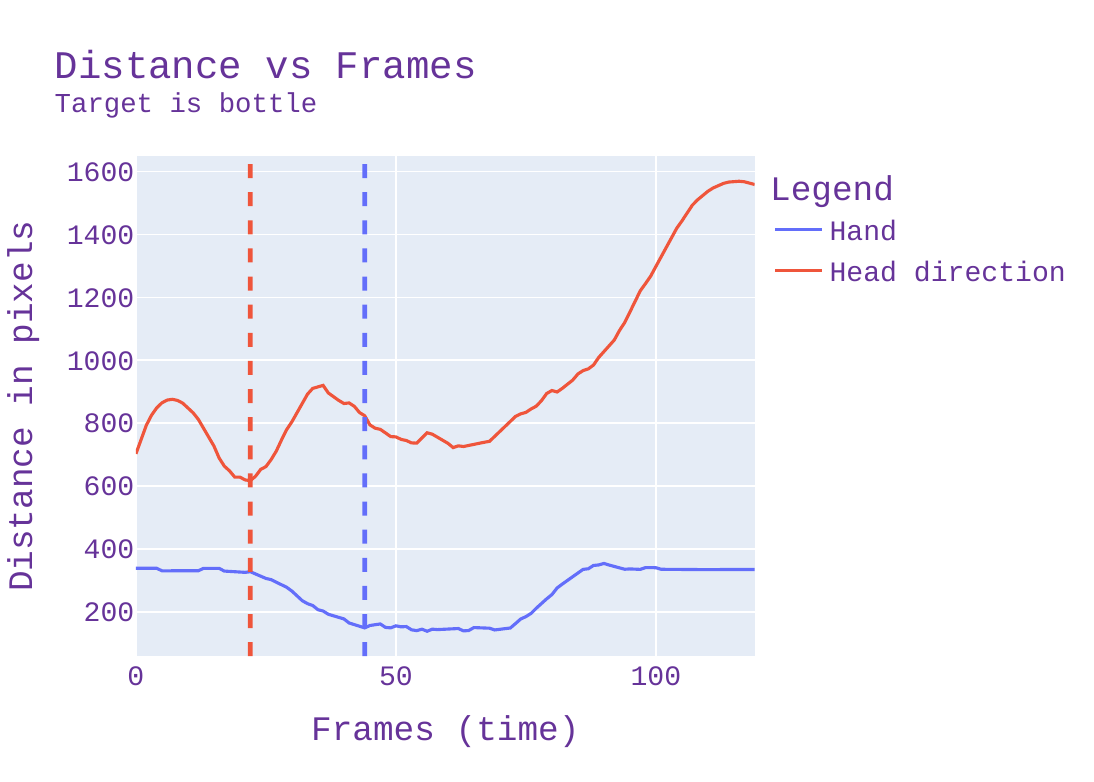}
  \caption{}
  \label{figSub_touch_bottle_2}
\end{subfigure}
\begin{subfigure}{.33\textwidth}
  \centering
  \includegraphics[width=.95\linewidth]{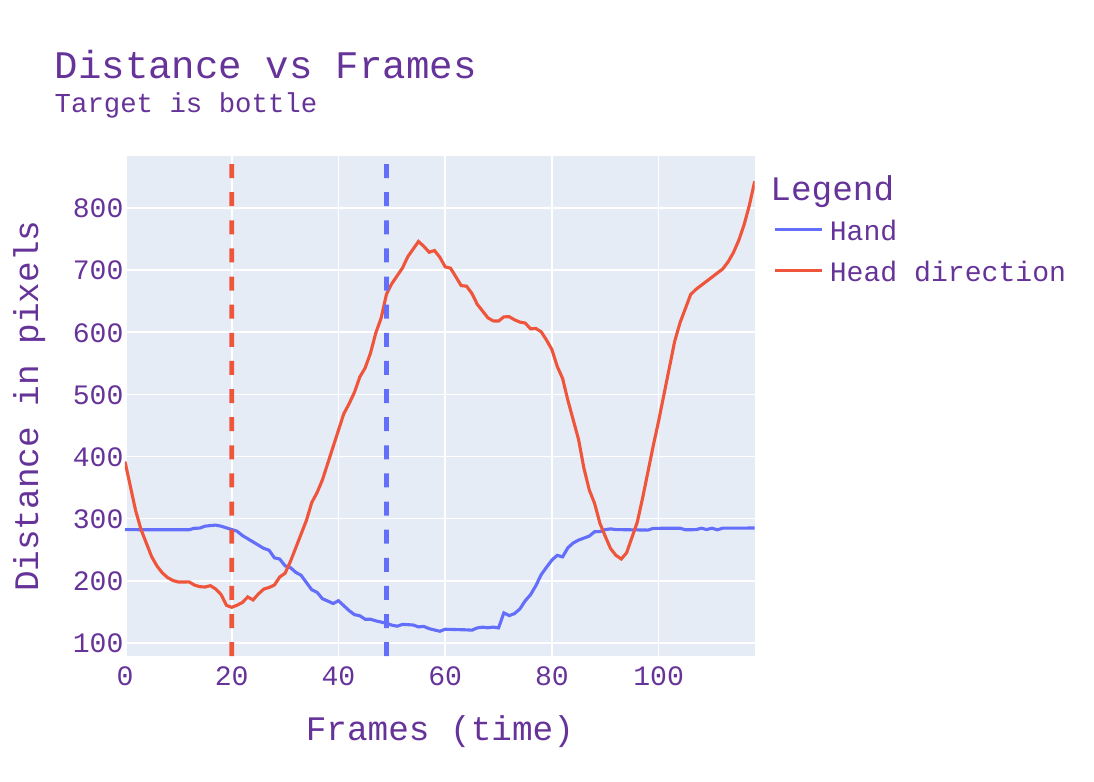}
  \caption{}
  \label{figSub_touch_bottle_3}
\end{subfigure}
\caption{Reaching the bottle in a \emph{touching action}.}
\label{fig_touch_bottle}
\end{figure*}

\begin{figure*}[p]
\begin{subfigure}{.33\textwidth}
  \centering
  \includegraphics[width=.95\linewidth]{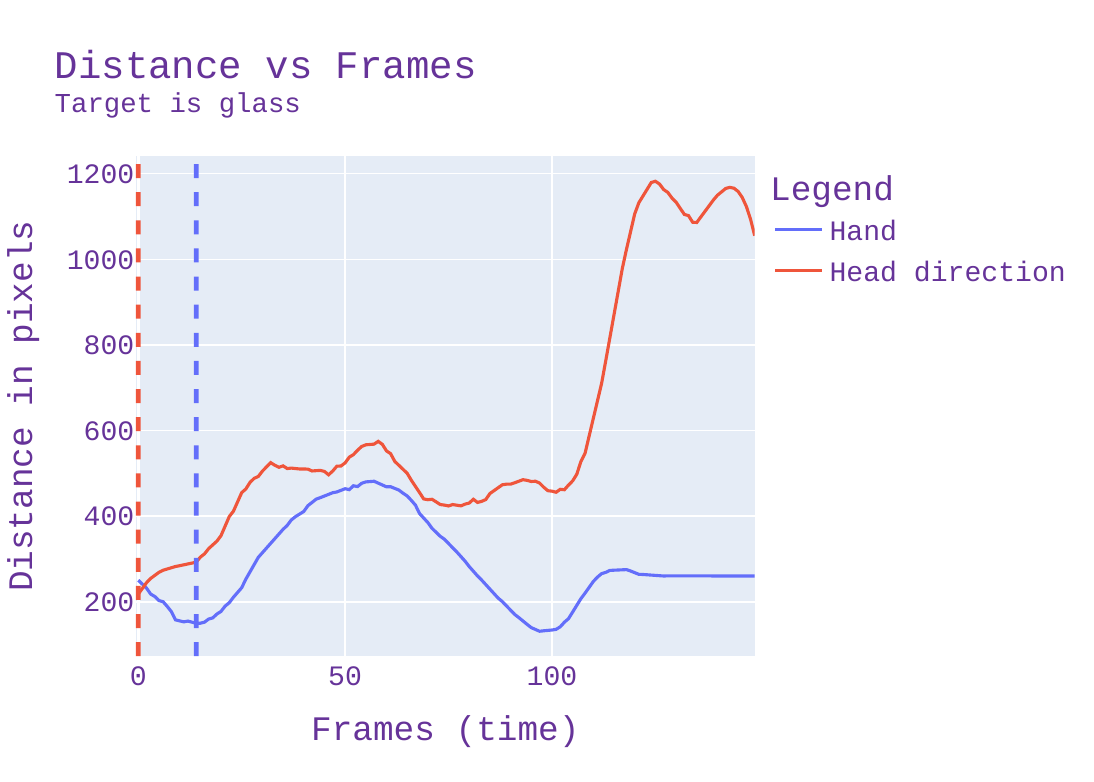}
  \caption{}
  \label{fig_reach_glass_1}
\end{subfigure}%
\begin{subfigure}{.33\textwidth}
  \centering
  \includegraphics[width=.95\linewidth]{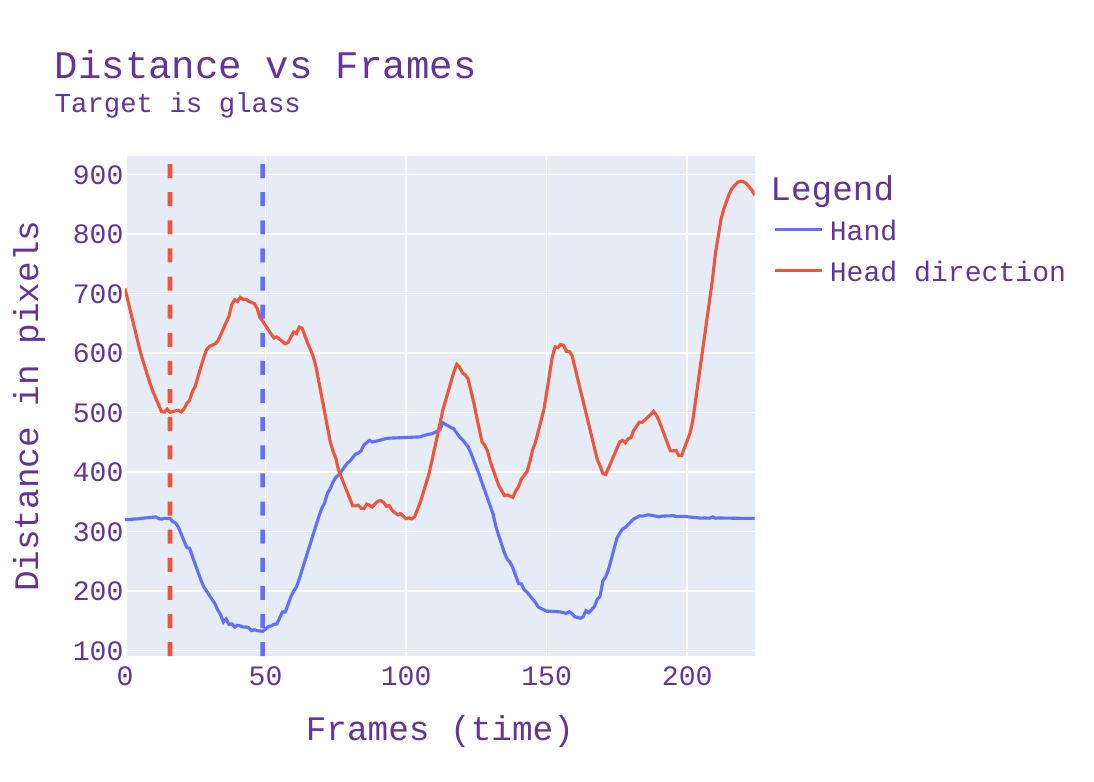}
  \caption{}
  \label{fig_reach_glass_2}
\end{subfigure}
\begin{subfigure}{.33\textwidth}
  \centering
  \includegraphics[width=.95\linewidth]{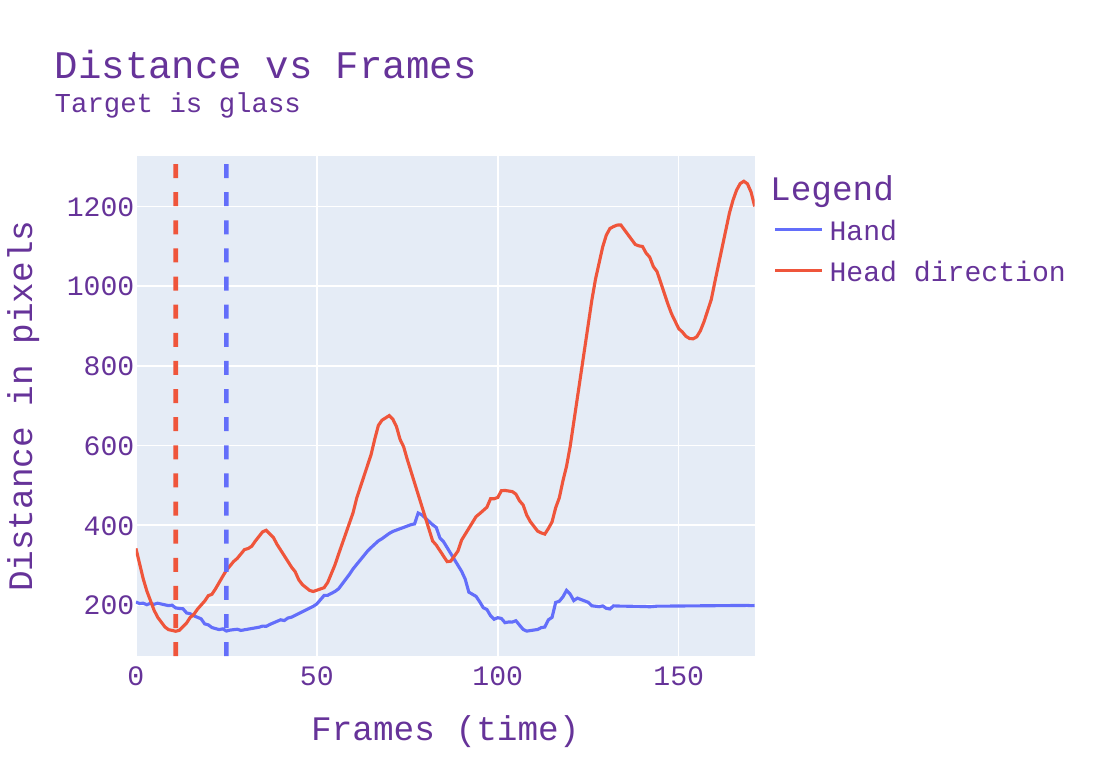}
  \caption{}
  \label{fig_reach_glass_3}
\end{subfigure}
\caption{Reaching the glass in the \emph{drinking action}. Which is in front of the subject, indeed in \ref{fig_reach_glass_1} the cup is in the line of sight of the subject from the very beginning of the action.}
\label{fig_graph_reach_glass}
\end{figure*}

\begin{figure*}[p]
\begin{subfigure}{.33\textwidth}
  \centering
  \includegraphics[width=.95\linewidth]{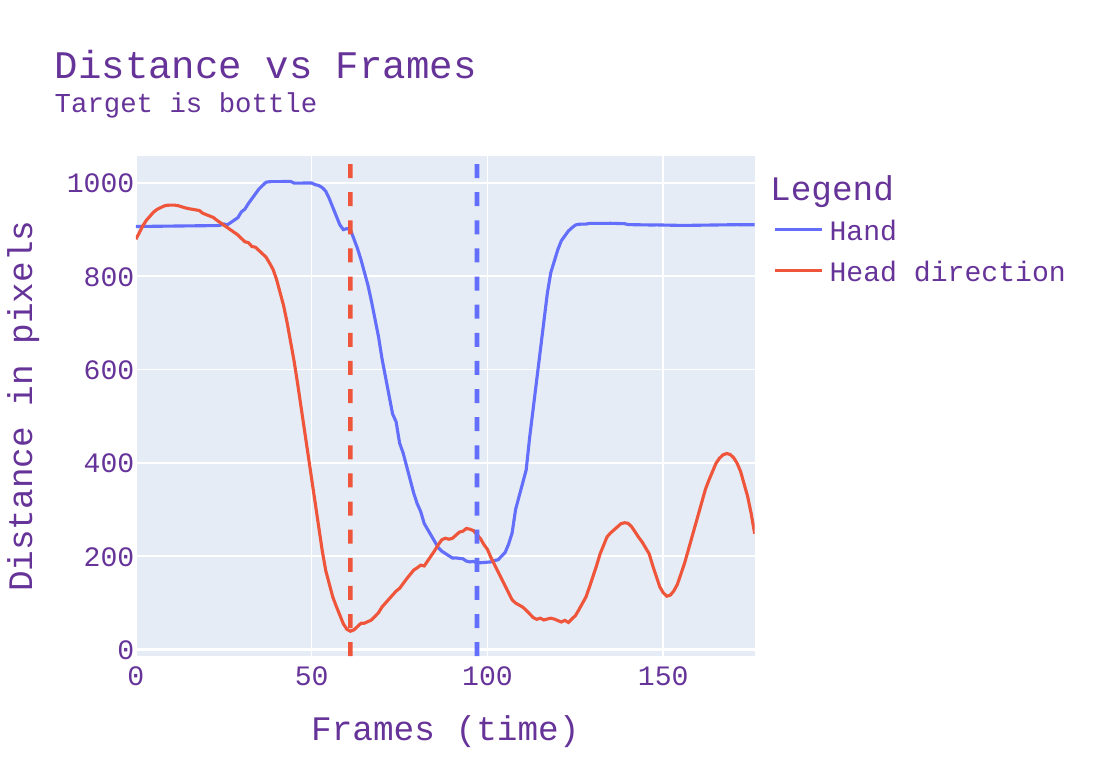}
  \caption{}
  \label{figSub_transport_1}
\end{subfigure}%
\begin{subfigure}{.33\textwidth}
  \centering
  \includegraphics[width=.95\linewidth]{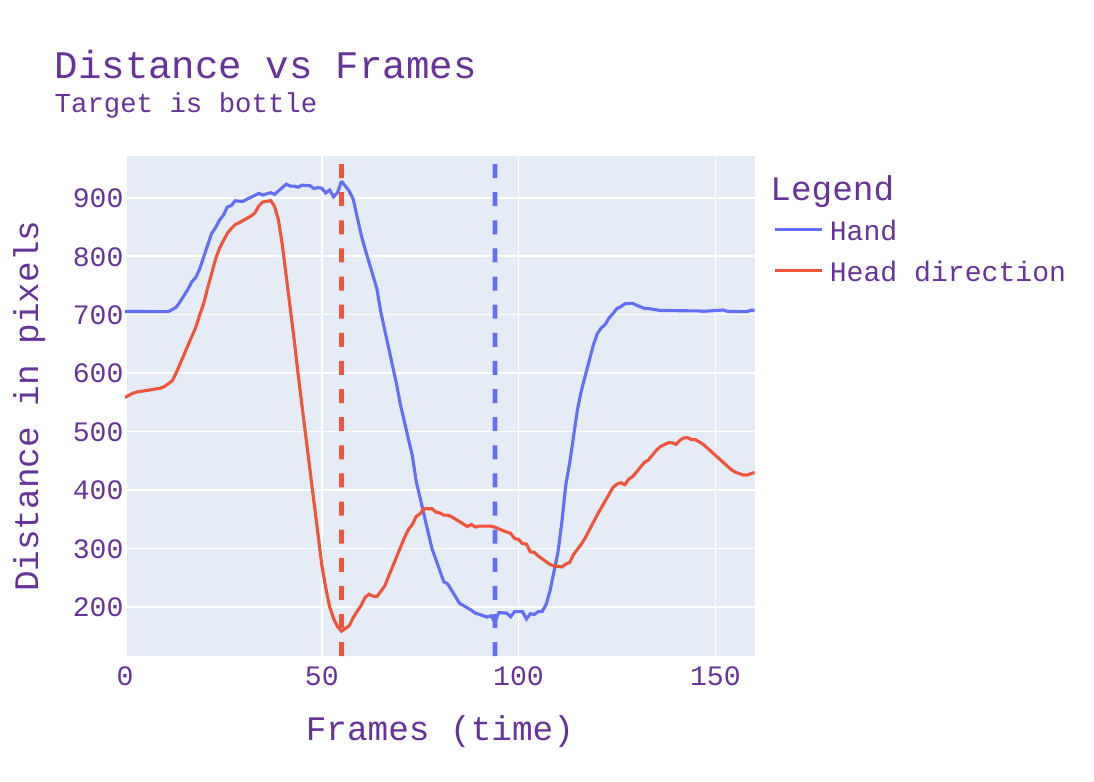}
  \caption{}
  \label{figSub_transport_2}
\end{subfigure}
\begin{subfigure}{.33\textwidth}
  \centering
  \includegraphics[width=.95\linewidth]{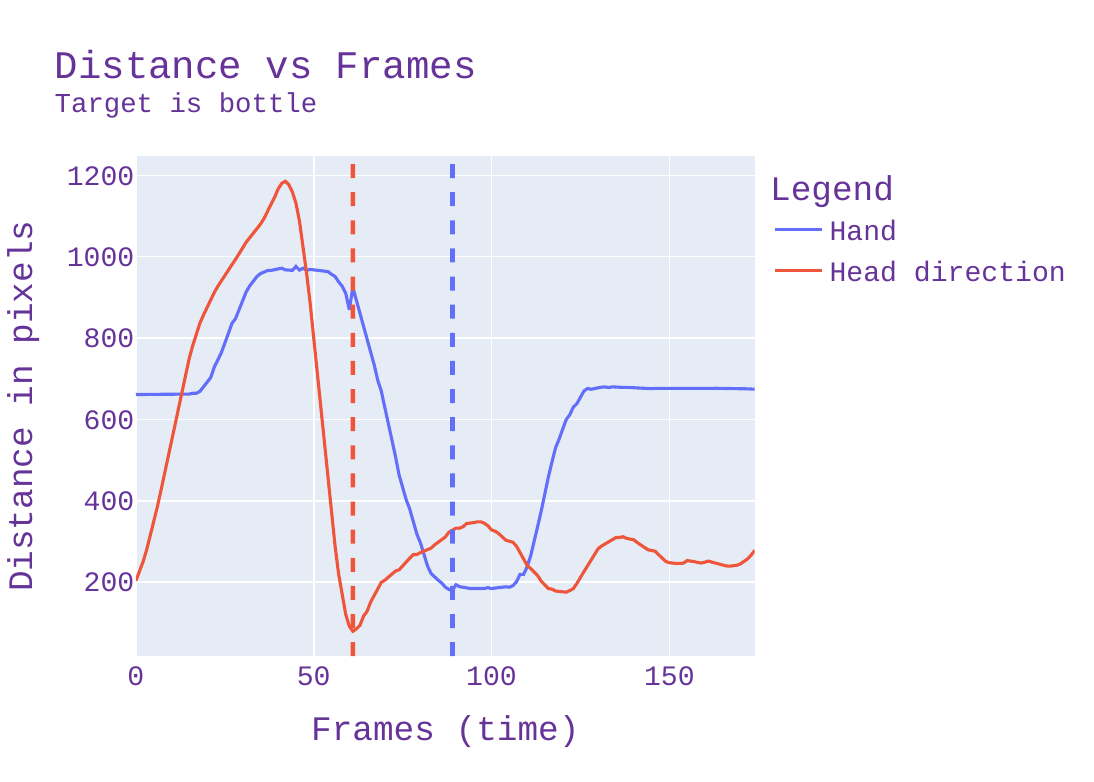}
  \caption{}
  \label{figSub_transport_3}
\end{subfigure}
\caption{Moving bottle to the target position in the \emph{transporting action}. It is very clear how the head goes to look toward that point and the hand follow consequently.}
\label{fig_graph_transport}
\end{figure*}

\section{Experimental Section}\label{sec_experiments}
\subsection{Dataset}
Stereo-HUM \cite{nicora2023gck} is an action classification dataset of 10 actions (drinking, eating crisps, opening and closing a bottle, playing with Rubik's cube, sanitise hands, touching a bottle, touching Rubik's cube, transporting a bottle, transporting a pen, transporting Rubik's cube) we acquired in-house. The full dataset is composed of 320 RGB videos of upper-body actions (a person seated at a table moving objects), each action is performed twice by each of the 16 participant subjects. \\
In this preliminary study, we selected 4 actions with characteristics in line with our needs for the analysis, in particular, the presence of a well-defined target, such as an object during a reaching or a destination of a transport action and the presence of the object in the COCO \cite{coco2014microsoft} classification labels. So we ended up using only four of the ten actions: drinking \emph{(\textbf{reaching the glass} $\rightarrow$ grasping it $\rightarrow$ transporting the glass to the mouth $\rightarrow$ drinking $\rightarrow$ putting back the glass on the table)}, touching the bottle \emph{(\textbf{reaching the bottle with the hand} $\rightarrow$ go back to the original position)}, transporting the bottle \emph{(\textbf{reaching the bottle} $\rightarrow$ grasping it $\rightarrow$ \textbf{moving it to the target position} $\rightarrow$ release it and go back to the original position)} and opening-closing a bottle \emph{(\textbf{reaching the bottle} $\rightarrow$ bringing the bottle closer to the person $\rightarrow$ opening the bottle $\rightarrow$ closing the bottle $\rightarrow$ put it back to the original position on the table and release it $\rightarrow$ put back the hand in the original position)}. Among the action phases, we highlighted in bold the ones we considered in our analysis.
The set of actions used contains 128 videos in total, each single-action set contains 32 videos.

\subsection{Experiments}
We present our experimental findings in two formats: first, a table detailing the anticipation times in seconds for each action type, averaged across all dataset examples; second, graphs that qualitatively illustrate the timing of gaze reaching the target relative to hand movement for selected actions.

Table \ref{tab_antipations} provides the average estimated \emph{anticipation} times. Negative values indicate that the head's projected position on the table reaches the target (either an object or a final position) before the hand does. All reported anticipation estimates are negative, indicating that the head consistently orients towards the target direction before the hand reaches the target position, which signifies the specific goal of the action.

The average anticipation is 15 frames, corresponding to 0.5 seconds since the videos have been acquired at 30fps. 

We present several results in Figures \ref{fig_transport_bottle}, \ref{fig_touch_bottle}, \ref{fig_graph_reach_glass}, and \ref{fig_graph_transport}, which illustrate the distances from the target for both gaze (in red) and hand (in blue). The vertical dashed lines indicate the minimum distance values used for computations: \emph{gazing\_target\_time}, \emph{touching\_object\_time} and \emph{target\_object\_time}. This minimum is not the absolute minimum of the sequence but the first minimum within a specified time window, calculated to encompass the initial part of the motion, where the focus is on \textit{reaching}, and a subsequent part, where the focus is on \textit{moving the object to the target position}.

Notably, the anticipation of the reaching target is consistent across different underlying actions or long-term goals. However, the amount of anticipation also depends on the object's position on the working table and its distance from the hand. In our scenario, when an object is positioned on one side of the table, such as the bottle in Figures \ref{fig_frames_example}, \ref{fig_transport_bottle}, and \ref{fig_touch_bottle}, the anticipation is pronounced and clearly observable. This is due to the relative positions of the hand, head, and target, which facilitate this observation. Conversely, when the object is directly in front of the person, such as the cup or glass, the anticipation is less evident because the head tends to remain in a neutral position, already close to the target. This is demonstrated in Figure \ref{fig_reach_glass_1}, where the minimum gaze-target distance occurs at the very beginning of the recording.

In the transport experiment depicted in Figures \ref{figSub_transport_frames1}, \ref{figSub_transport_frames2} and \ref{figSub_transport_frames3} the measured anticipation focuses on the object's final position or the action's ultimate goal. This is clearly illustrated in the plots of Figure \ref{fig_graph_transport} where it can be observed that the hand consistently follows the head. Across all three examples shown, the degree of anticipation is remarkably similar.

\begin{table}[]
    \centering
    \begin{tabular}{lcccc}
    \toprule
        original action & measured quantity & mean [s] & std [s] & median [s]\\
        \midrule
        transport bottle & reach bottle & -0.51 & 0.35 & -0.43\\
        touch bottle & reach bottle & -0.64 & 0.34 & -0.63\\
        open-close bottle & reach bottle & -0.54 & 0.35 & -0.50\\
        drinking & reach glass & -0.49 & 0.85 & -0.77\\
        \midrule
        transport bottle & object to target & -0.70 & 0.52 & -0.78\\
        \bottomrule
    \end{tabular}
    \caption{Anticipation in seconds using the head-pose projection with respect to the hand movement. The first column exposes the action class recorded in the dataset, whereas the second column shows the measured quantity for this assessment. The measures are in seconds (the minus is to indicate anticipation of the head with respect to the hand).}
    \label{tab_antipations}
\end{table}
\section{Conclusion} \label{sec_conclusion}
We presented a preliminary experiment to show that visual cues derived from the head direction can be profitably used to efficiently anticipate the action's goal of reaching and transporting movements in controlled interaction scenarios.
This analysis lays the foundation for further exploration and development of a predictive model that could be integrated into a robotic system to enhance its social interaction abilities and its understanding of nonverbal information. In group-robot interaction, for instance, the robot might benefit from the proposed analysis for 
predicting to whom a person will pass the object in a group, or how the focus of attention of the group members evolves over time. In this sense, it is worth noticing that our method can also work online by design.\\
Our current investigation aims at a more comprehensive assessment of the method.

\section*{Acknowledgments}
This work has been carried out at the Machine Learning Genoa (MaLGa) center, Universit\'a di Genova (IT). It has been
funded by the European Union—NextGenerationEU and by the Ministry of University and Research (MUR), National Recovery and Resilience Plan (NRRP), project “RAISE—Robotics and AI for Socio-economic Empowerment” (ECS00000035).

\bibliographystyle{plainnat}
\bibliography{references}

\end{document}